# Detecting Emerging Technologies and their Evolution using Deep Learning and Weak Signal Analysis

Ashkan Ebadi[1,*], Alain Auger[2], and Yvan Gauthier[3]

1. National Research Council Canada, Montreal, QC H3T 1J4, Canada
2. Science and Technology Foresight and Risk Assessment Unit, Defence Research and Development Canada, Ottawa, Ontario, Canada
3. National Research Council Canada, Ottawa, ON K1K 2E1, Canada
[*] Corresponding author: Ashkan Ebadi (ashkan.ebadi@nrc-cnrc.gc.ca)

**Abstract**
Emerging technologies can have major economic impacts and affect strategic stability. Yet, early identification of emerging technologies remains challenging. In order to identify emerging technologies in a timely and reliable manner, a comprehensive examination of relevant scientific and technological (S&T) trends and their related references is required. This examination is generally done by domain experts and requires significant amounts of time and effort to gain insights. The use of domain experts to identify emerging technologies from S&T trends may limit the capacity to analyse large volumes of information and introduce subjectivity in the assessments. Decision support systems are required to provide accurate and reliable evidence-based indicators through constant and continuous monitoring of the environment and help identify signals of emerging technologies that could alter security and economic prosperity. For example, the research field of hypersonics has recently witnessed several advancements having profound technological, commercial, and national security implications. In this work, we present a multi-layer quantitative approach able to identify future signs from scientific publications on hypersonics by leveraging deep learning and weak signal analysis. The proposed framework can help strategic planners and domain experts better identify and monitor emerging technology trends.

# 1. Introduction

The complex and rapidly evolving nature of modern science has given rise to emerging technologies with transformative and disruptive characteristics, as recently exemplified by deep learning [1], cryptocurrencies [2], mRNA vaccines [3], and solar cells [4]. Such technologies have changed the landscape of existing industries while creating new economic opportunities and affecting societies and the lives of people [5]. Given the highly competitive and evolving environment, planners always seek insights into emerging technologies that could affect long-term strategic stability [6], economic development, and national security.

Early detection of new technology trends is of critical importance for governments and businesses, as it enables them to identify opportunities and risks quickly and react to them accordingly by formulating appropriate research, development, and innovation strategies [7]. Several Scientometrics studies have considered various data sources, such as scientific publications and patents, and employed different techniques, such as bibliometrics and keyword network analysis, to identify the emergence and evolution of new technologies [8, 9, 10]. Using co-word analysis and by means of scientometric indicators, Lee [11] identified emerging research themes in the field of information security by analyzing patterns and trends.

Abercrombie et al. [12] analyzed the relationships among several data sources and proposed a scientometric model to track the emergence of new technologies. In another study, Kim and Zhu [13] applied dynamic topic modeling to identify and investigate the thematic patterns and emerging trends of the scientific publications in scientometrics. More recently, Park and Yoon [14] presented a quantitative approach to discover potential future technological opportunities from the patent-citation network.

Recent advances in digital technology have made such data more accessible than ever. Various science and technology (S&T) databases exist for many fields of innovation, with data continuously increasing in size and variety [15]. On the other hand, the fast-evolving nature of modern science and technology has made strategic planning more complex [16]. The incomplete and asynchronous information, in some cases, adds to this complexity [17]. These attributes along with the availability of advanced computer science algorithms provide new opportunities to gain better insights from S&T databases. For example, in a recent study, Xu et al. [18] combined several machine learning models and proposed a framework to identify and foresight emerging research topics at the thematic level. Computerized systems can help human experts keep pace with increasing and evolving data and extract knowledge about a specific technology domain [19]. For instance, the research field of hypersonics is rapidly progressing and has various commercial and military applications [20]. The term hypersonic means "pertaining to or moving at a speed greatly in excess of the speed of sound, usually meaning greater than Mach 5. All speeds in excess of the speed of sound are supersonic, but to be hypersonic requires even higher speed" [21]. Research in hypersonics is now central to many civilian and military aerospace programs.

In this research, we propose a multi-layer approach to extract signals and trace their evolution, combining various techniques such as natural language processing (NLP), deep learning, and weak signal analysis. Our contribution is threefold: first, we improve upon keyword extraction by going beyond traditional statistical approaches and leverage a pre-trained transformer-based deep learning technique, namely BERT [22], and get domain experts to validate the extracted keywords. Second, we apply recent concepts of weak signal analysis to the early detection of technology emergence. Third, we offer an end-to-end, modular approach that can be re-applied to other technology fields and leverage additional data sources. The resulting approach enables faster comprehensive and large-scale quantitative analyses that can complement expert-based methods. As a test case to measure the ability of the approach to identify trends and extract early signals, we focus on scientific publications in the domain of hypersonics for the period from 1985 to 2020.

The remainder of the paper is organized as follows: previous works on text mining and weak signal analysis are reviewed in Section 2. Section 3 describes data and methodology. We present our findings in Section 4 and discuss them in Section 5. Section 6 covers the limitations of this work and draws some directions for future research.

## 2. Related Work

Our proposed approach is based on automatic keyword extraction using state-of-the-art techniques and weak signal analysis. In this section, these topics are briefly introduced and previous work is discussed.

### 2.1. Automatic Keyword Extraction

Automatic keyword extraction (AKE) is the process of automatically extracting representative keywords (and/or phrases) from a document (or a collection of documents) [23]. The extracted keywords may help the user to grasp a high-level understanding of the content presented in the respective documents. This is crucial especially if the user deals with large data sets, which is usually the case in real-world applications. AKE has been widely studied. At a high-level, keyword extraction approaches can be classified into four categories: 1) statistical approaches, 2) linguistics approaches, 3) machine/deep learning approaches, and 4) hybrid methods [24]. We briefly introduce these approaches in the following sections.

### 2.1.1. Statistical Approaches

Statistical methods for keyword extraction are mostly based on simple statistics calculated from non-linguistic features of a (set of) document(s) [25]. Several papers have reported results for the extraction of a set of keywords from a corpus using simple statistical measures such as term frequency (TF) [26], word co-occurrences [27], and term frequency-inverse document frequency (TF-IDF) [28]. Since these approaches are based on the frequency of terms/occurrences, their results could be noisy and not very precise [25].

### 2.1.2. Linguistic Approaches

Linguistic methods use natural language processing techniques and linguistic features of words in a (set of) document(s) to detect and extract representative keywords [25]. For example, [29] used lexical analysis to summarize documents, [30] employed syntactic analysis to improve automatic keyword extraction, and [31] performed discourse analysis to automatically structure and summarize documents.

### 2.1.3. Machine Learning Approaches

Machine learning techniques are used to extract keywords from the corpus. The learning phase can be done in a supervised setting where a model is trained using (large) annotated text corpora or in an unsupervised setting without using any annotated dataset. There are several works in the literature that performed supervised learning approaches for keyword extraction using various machine learning techniques such as support vector machine (SVM) [32] and Naive Bayes [33]. Rapid Automatic Keyword Extraction (RAKE) [34] is an example of an unsupervised, domain- and language-independent method for automatic keyword extraction from documents. YAKE! [35] is another example of an unsupervised automatic keyword extraction approach using statistical text features. With recent advancements in language representation such as the Bidirectional Encoder Representations from Transformers (BERT) method [22], new techniques are introduced that use BERT embeddings for keyword and keyphrase retrieval [36, 37].

### 2.1.4. Hybrid Approaches

Hybrid methods combine the above-mentioned approaches or use heuristics to calculate the best features from the target text corpora and use them to extract keywords and/or keyphrases [38, 39].

### 2.2. Weak Signal Analysis

Detection of weak signals, trends, and issues in the evolving technology landscape as early as possible has been emphasized in the literature for strategic planning [40, 41]. Different types of changes could be of interest from an unexpected discontinuity in a technology trend to a new emerging (mega-)trend with a high impact on society and the technology landscape itself [42]. The weak signal concept was first introduced by Ansoff [40] as an alternative to strategic planning in the 1970s and 1980s. Later in 1997, Coffman [43] provided a more detailed definition of Ansoff's, defining a weak signal as a signal that: 1) can affect the business or its environment, 2) is new to the receiver of the signal despite not being necessarily new to others, 3) could be difficult to track, 4) can be regarded as a threat or opportunity to a specific group, 5) can be downgraded by those who perceive and know this signal, 6) needs time to become mature and mainstream, and 7) is an opportunity to learn and grow. According to Saritas [44], weak signals are "early signs of possible but not confirmed changes that may later become more significant indicators of critical forces", hence, they can be regarded as signals about future trends of a given technology. In other words, these signals are potential indicators of the discontinuity or emergence of a target technology even with no current significant impact [45]. More recently in 2021, van Veer and Ortt [46] reviewed 68 definitions of weak signals and proposed a common unified definition that crosses foresight domains: "A perception of strategic phenomena detected in the environment or created during interpretation that are distant to the perceiver's frame of reference".

Weak signal analysis is one of the methods used to detect emerging terms in a specific technology domain. According to Hiltunen [47], future signs are current oddities and strange issues that are key indicators of predicting future changes. Hiltunen [47] suggested three dimensions for future signs: 1) signal, i.e., an indicator of visibility, 2) issue, i.e., an indicator of diffusion of a future sign, and 3) interpretation, i.e., the meaning of the future sign to the receiver. Signs can be categorized into weak to strong signals in this three dimensional space. Figure 1 shows Hiltunen's three dimensions of future signs and the characteristics of weak and strong signals [47]. In this future sign space, a weak signal has low levels of signal, issue, and interpretation and can turn into a strong signal exhibiting higher levels of the mentioned dimensions.

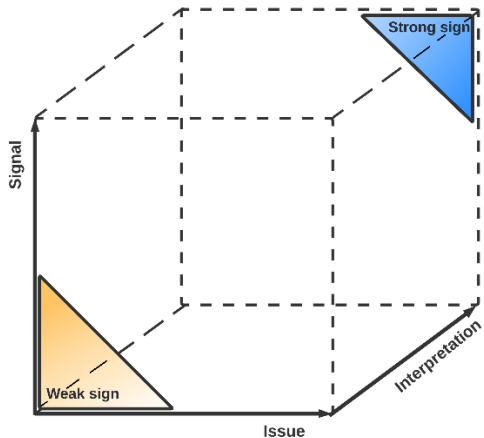

**Figure 1.** Weak and strong signals in Hiltunen's three dimensions of future signs.

Several qualitative studies have been conducted on the concept of the weak signals to analyze business environments, assist corporate decision-making, and strategy-creation process [48, 49], support strategic planning and foresight services in corporations [50], and utilize an image-based medium to trigger employees' future thinking in analyzing organizations [51], among others. Although Hiltunen's qualitative framework could assist experts in identifying signals, it may suffer from subjectivity and may require lots of resources to be implemented [52].

Yoon [45] proposed a quantitative method for Hiltunen's framework [47] based on text mining to mitigate analysts' subjectivity and overcome the limitations of the qualitative approach. Despite its limitations, term frequency is often considered as a measure of the importance of a term [53]. Document frequency represents the number of documents in which a specific term has appeared [54] and is used as a measure of dissemination of a term in a collection of documents [53]. Using term and document frequencies, Yoon [45] described a weak signal as a term with low term and document frequencies and a high growth rate. Yoon [45] also described a strong signal as a term with high term and document frequencies and a high growth rate.

# 3. Data and Methodology

## 3.1. Data

Researchers mostly convey their scientific discoveries and findings to the scientific community and the general public via scientific publications [55], hence it is considered the major output of scientific research [56]. Therefore, in this study, we focused on scientific publications as the primary data source to identify emerging technologies in hypersonics using machine learning and weak signal analysis.

Data collection and preparation involved several steps. First, using the string "hypersonic*" as the search query, we collected the bibliographic data of hypersonics-related publications that were published within the period from 1985 to 2020 from the Elsevier's Scopus database. A total of (n = 21, 669) articles was retrieved in August 2021. This included many meta-data about each of the retrieved articles such as date of publication, title, abstract, and list of authors and affiliations. We checked and ensured that for all of the collected articles either the title or abstract was available. We then created a new field combining title and abstract and removed duplicated records (n = 79). The final dataset contains (n = 21, 590) publication records spanning from 1985 to 2020.

Figure 2 shows the annual distribution of publications over the examined period. As shown below, there is an overall growth in the number of publications. The number of publications has almost plateaued between 2017 and 2020, after 15 years of sustained growth. The majority of the publications were either published in scientific journals or presented at conferences.

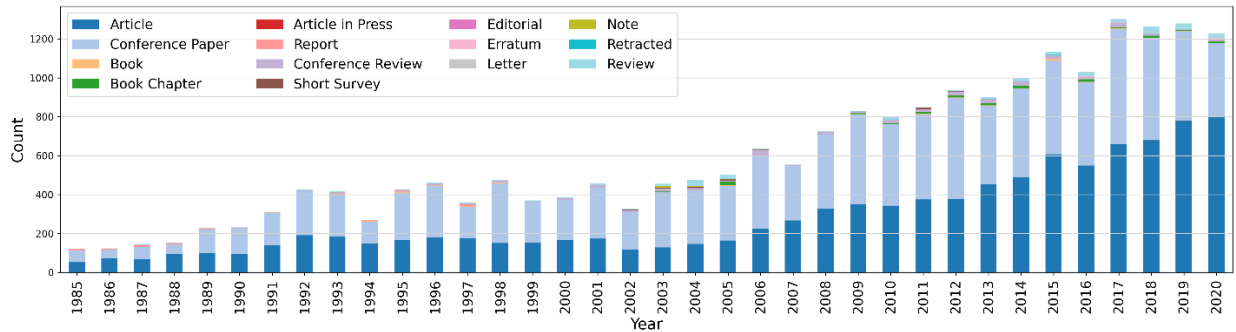

**Figure 2.** Annual distribution of the publications in the dataset (n = 21, 590).

## 3.2. Methodology

Figure 3 shows the high-level conceptual flow of the analytic pipeline used in this study. After data collection, the process is followed by preparing the collected data and extracting keywords. Next, keyword portfolio maps are constructed and signals are extracted. Then, the extracted signals are post-processed and verified by domain experts, and are categorized into defined categories. Finally, the strength of the extracted signals as well as the evolution of the categories over time are analyzed and depicted. The rest of this section describes our methodology and the analytic pipeline in detail.

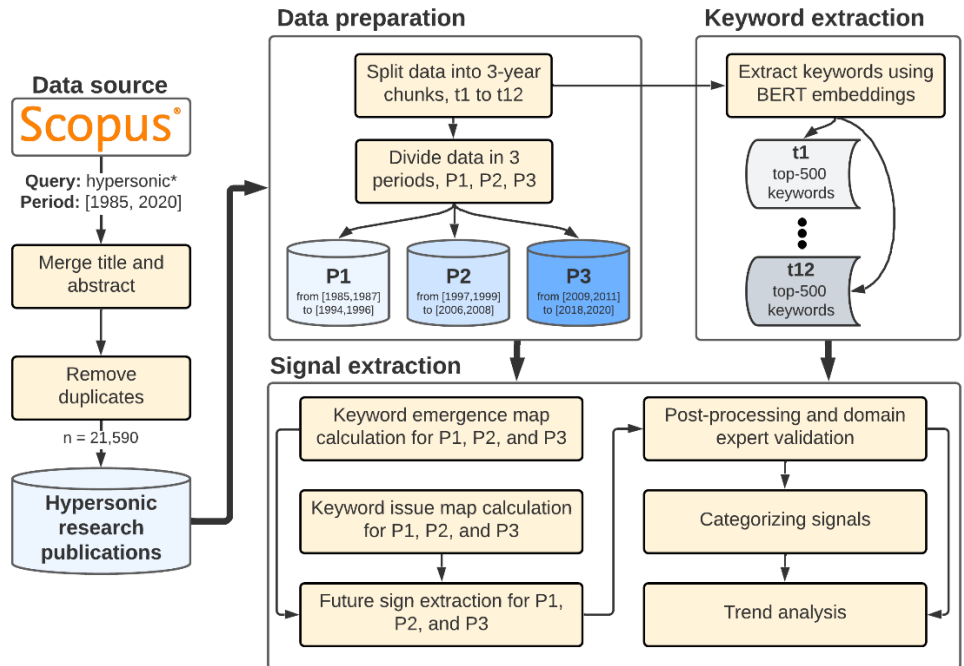

**Figure 3.** The high-level conceptual flow of the analyses.

### 3.2.1. Data Intervals

As mentioned in Section 3.1, our dataset covers hypersonics publications within the period from 1985 to 2020. We experimented with various time intervals for dividing the examined period and found that a 3-year window would produce the best results in terms of a representative sampling of the keywords across the entire dataset. Therefore, we divided the examined time period into 3-year time intervals, starting from the [1985, 1987] period to [2018, 2020]. We will refer to these intervals as t1 to t12 in the rest of the paper. Based on the created 3-year data intervals, we divided the dataset into 3 periods as follows:

- **Period 1 ($P_1$):** data from [1985, 1987] to [1994, 1996], i.e., $t_1$ to $t_4$.
- **Period 2 ($P_2$):** data from [1997, 1999] to [2006, 2008], i.e., $t_5$ to $t_8$.
- **Period 3 ($P_3$):** data from [2009, 2011] to [2018, 2020], i.e., $t_9$ to $t_{12}$.

### 3.2.2. Keyword Extraction

Several approaches have been proposed in the literature to extract keywords from a corpus. These approaches include but are not limited to information filtering [57], graph-based methods [58], centrality measures [59], and domain knowledge [60]. These techniques are mostly based on term frequencies in the corpus. Over the recent years, the natural language processing (NLP) community has developed and used large language models (LLMs) such as Bidirectional Encoder Representations from Transformers (BERT) [22]. LLMs are self-supervised task-agnostic neural networks and are good at learning general linguistic patterns in a corpus. This property has boosted the performance of downstream models by focusing on specific tasks rather than learning linguistic patterns [61].

In this study, we used the BERT language model [22] to extract keywords. We applied BERT to extract document-level representation for each of the time intervals defined in Section 3.2.1, i.e., $t_1$ to $t_{12}$. Word embeddings were extracted for n-grams and cosine similarity was used to find the most representative keywords (and phrases) of each document in the dataset. Following these steps, 500 keywords were extracted for each of the 12 time intervals, resulting in 6,000 keywords in total. We then combined these

keywords in a list and removed duplicates. Finally, we applied a customized stop words list to remove common words. The final keyword set contained 4,366 terms.

### 3.2.3. Keyword Emergence Map

The frequency of a keyword in a time period may reflect its degree of visibility during that period. It is assumed that highly frequent keywords with more occurrences in recent years are important indicators of emergence. Term frequencies were used to calculate keywords' degree of visibility. Following Yoon [45], we used a time-weighted formula to calculate the degree of visibility (DoV) of the extracted keywords (n = 4,366) in each of the 12 aforementioned time intervals, i.e., $t_1$ to $t_{12}$:

$$DoV_{ij} = \left(\frac{TF_{ij}}{N_j}\right) * \left(1 - w * (n - j)\right) \qquad (1)$$

where $TF_{ij}$ is the total frequency of keyword $i$ in time interval $j$, $N_j$ is the total number of documents in time interval $j$, $n$ is the number of time intervals (n = 12, in our experiment), and $w$ is a constant time-weight. We set w = 0.05 as optimal after experimenting with different values and carefully reviewing the results. The Keyword Emergence Map (KEM) can next be created from the calculated DoV values. In KEM, the x-axis represents the average term frequencies of the keywords, and the y-axis indicates the DoV's growth rate, calculated as the geometric mean. The quadrants in the KEM are divided based on the medians of the respective values on each axis. This creates 4 quadrants in the KEM map as follows: 1) the top-right region represents strong signals that are keywords with high average term frequency and high average DoV growth rate, 2) the top-left region represents weak signals that are keywords with low average term frequency and high average DoV growth rate, 3) keywords in the bottom left region are marked as latent signals, and 4) bottom right of the KEM represents well-known but not strong signals that are keywords with high term frequency but low DoV growth rate. Using these definitions, we assigned a signal label to each of the keywords in KEM in each examined period, i.e., $P_1$ to $P_3$.

### 3.2.4. Keyword Issue Map

In Hiltunen's model [47], the issue dimension focuses on how much the signals are disseminated. To measure the diffusion rate quantitatively, we followed Yoon's approach [45] and used the document frequencies of the extracted keywords. For this purpose, we first used the following time-weighted formula to calculate the degree of diffusion (DoD) of the extracted keywords in each of the 12 aforementioned time intervals, i.e., $t_1$ to $t_{12}$:

$$DoD_{ij} = \left(\frac{DF_{ij}}{N_j}\right) * \left(1 - w * (n - j)\right) \qquad (2)$$

In equation (2), $DF_{ij}$ is the document frequency of keyword $i$ in time interval $j$, $N_j$ is the total number of publications in time interval $j$, $n$ is the number of time intervals (n = 12, in our experiment), and $w$ is a constant time-weight, w = 0.05. Next, the Keyword Issue Map (KIM) was generated by plotting the average time-weighted increasing rate of keywords' document frequencies on the y-axis, and keywords' average document frequencies on the x-axis. According to Yoon [45], weak signals correspond to those terms with relatively low document frequency but a high growth rate. Therefore, similar to the approach explained in the previous section, we divided the KIM into 4 quadrants based on the medians of the values on each axis, as follows: 1) the top-right region represents strong signals that are keywords with high average document frequency and high average DoD growth rate, 2) the top-left region represents weak signals that are terms with low average document frequency and high average DoD growth rate, 3) keywords in the bottom-left region are labeled as latent signals, and 4) bottom-right of the KIM represents well-known but not strong signals. Using these definitions, we assigned a signal label to each of the keywords in KIM in each examined period, i.e., P1 to P3.

### 3.2.5. Signals Extraction

Figure 4 shows the process of extracting signals from KEMs and KIMs by analyzing their intersection. Signals appearing in the same quadrant in both KEM and KIM indicated similar degrees of visibility and dissemination and were extracted. We did this process for KEMs and KIMs in each of the examined periods, i.e., P1 to P3, and extracted the final weak, strong, and well-known but not strong signals. For example, in Figure 4, "Term-A" and "Term-B" are extracted as weak signals since they are both listed as weak signals in both maps. However, "Term-C" is not listed as a weak signal because it does not appear as a weak signal in both maps. The extracted signals were then verified and validated by domain experts.

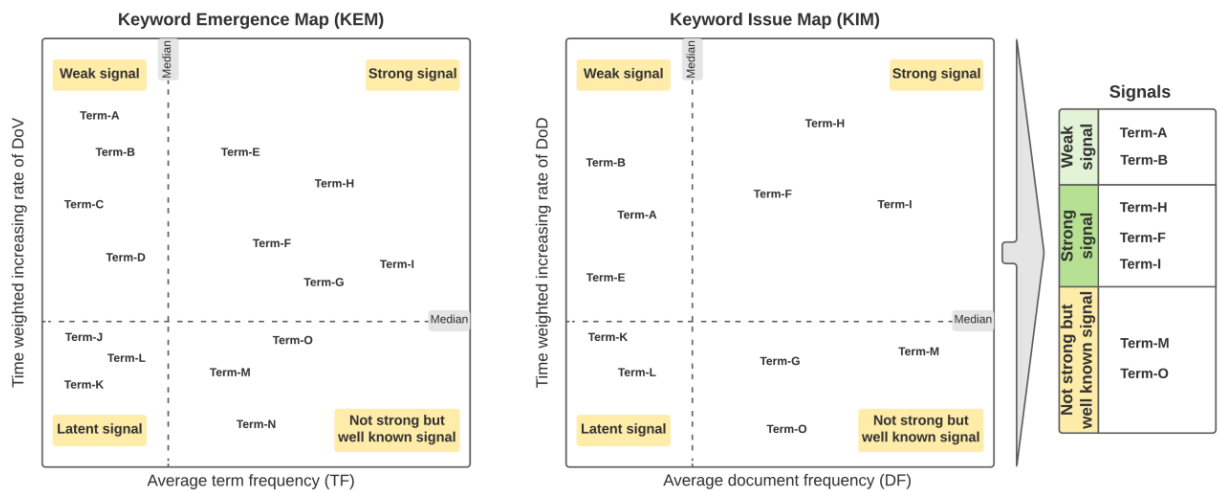

**Figure 4.** The process of extracting signals from KEM and KIM.

## 4. Results

### 4.1. Extracted Terms Categories

Following the methodology that was explained in Section 3.2, senior scientists with domain expertise in hypersonics carefully verified and validated the extracted signals (n = 442) and classified them into 5 high-level categories. Table 1 shows the defined categories along with a set of sample terms listed under each category. For example, terms such as "aeroshell", "aerobrake", and "aerohydrodynamic" were grouped into the Aerothermodynamics category. Letters in brackets, listed under the Abbreviation column of Table 1, are used in the rest of this paper to refer to term categories. As illustrated, the *Materials and structures* and the *Guidance, navigation, and control* categories contain the highest and the lowest number of terms, respectively.

**Table 1.** The defined high-level categories of the extracted signals along with a sample set of terms for each category.

| Category | Abbreviation | Sample keywords (n = 10) |
|:---:|:---:|:---:|
| Aerothermodynamics (n = 103) | [a] | aeroshell, aerobrake, aerohydrodynamic, aeromechanics, aerosol, aerospike, airfoil, anisotropic, blast, buoyancy, ... |
| Guidance, navigation, and control (n = 54) | [g] | accelerometer, aerodisk, ailerons, altimeter, azimuthal, backscattered, controllability, dynamic surface control, |
| | | electronic flight display, extender, ... |
| Materials and structures (n = 133) | [m] | ablation, adiabatic, aeroelastic, airbag, airframes, alkanes, alloying, anharmonicity, antiwindup, argon, ... |
| Modelling, simulation, and analysis (n = 59) | [s] | astrophysics, bellman, bhatnagar, brownian, burnett, cad, captive, chapman, dirac, discretisation, ... |
| Vehicles, propulsion, and fuels (n = 93) | [v] | acetylene, aegis, afterburn, ammonia, apertures, air turbo rocket, autoignition, ballistic, bursts, capsule, ... |

Figure 5 shows the coverage percentage of defined categories in each examined period. As shown in the figure, the "Materials and structures" category has a higher coverage in all periods, except for the strong signals in $P_1$ where it is dominated by the "Vehicles, propulsion, and fuels" category. The proportion of categories calculated based on all the extracted terms is almost similar in $P_1$, $P_2$, and $P_3$, however, more changes in terms of categories coverage is observed for strong signals. Moreover, although the proportion of the "Modelling, simulation, and analysis" category is not high compared to other categories, its weak signals coverage has increased from $P_1$ to $P_3$.

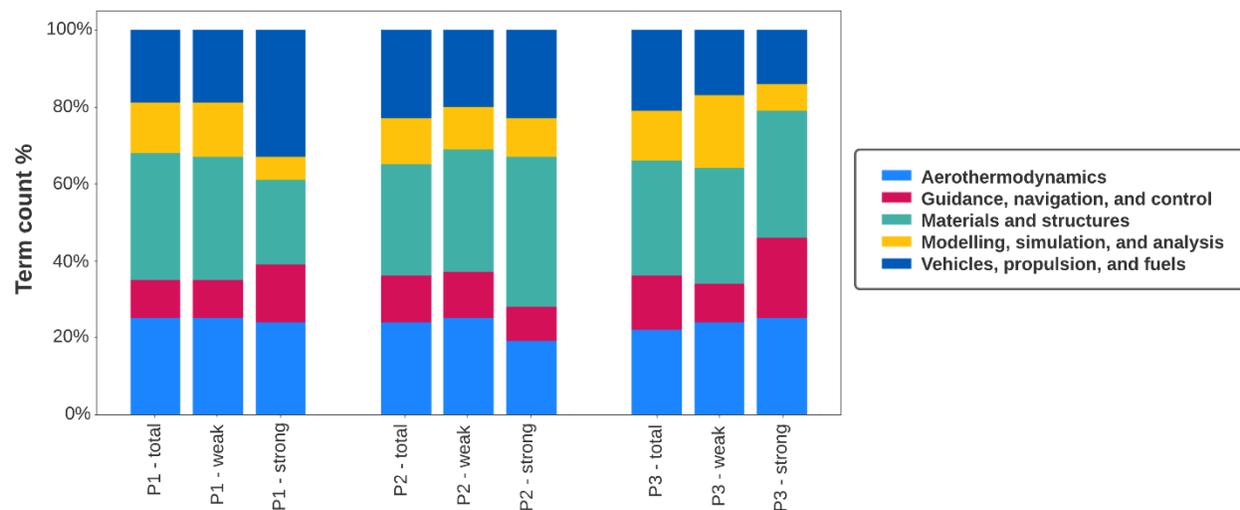

**Figure 5.** Percentage of related terms per category in each period for all terms, weak signals, and strong signals.

### 4.2. Signals Evolution over Time

Before analyzing signals' temporal changes, we first investigate the existence of temporal evolution in the documents during the examined period. For this purpose, using the extracted keywords for each examined time interval (n = 500 ∗ 12, as explained in Section 3.2.2), a bipartite graph G(V,E) of keywords and time intervals is created such that each time interval is linked to its representative keywords. Next, the Louvain modularity method [62] is applied to the graph to detect the community structure. Figure 6-a shows the generated <time interval, keyword> graph. The bipartite graph contains two types of nodes: 1) time intervals that are the bigger nodes in the graph, e.g., [1985, 1987], and 2) keywords that are the smaller nodes in the graph, e.g., "radar". Node colors in the figure represent different communities where, for example, keywords of [2012, 2014] and [2015, 2017] belong to the same community. Figure 6-b shows the degree distribution of keyword nodes. As shown in the figure, most of the nodes are of degree 1 indicating that they have only appeared in 1 time interval, therefore, the extracted keywords are highly specific. Figure 6-c shows the same graph as in Figure 6-a except for the nodes with a degree of 1 are excluded. As it is observed in Figures 6-a and –c the community structure implies the existence of a temporal trend/evolution since there are 8 different groups (communities) of time intervals identified. This confirms that there is a relationship between the extracted keywords and the publication dates.

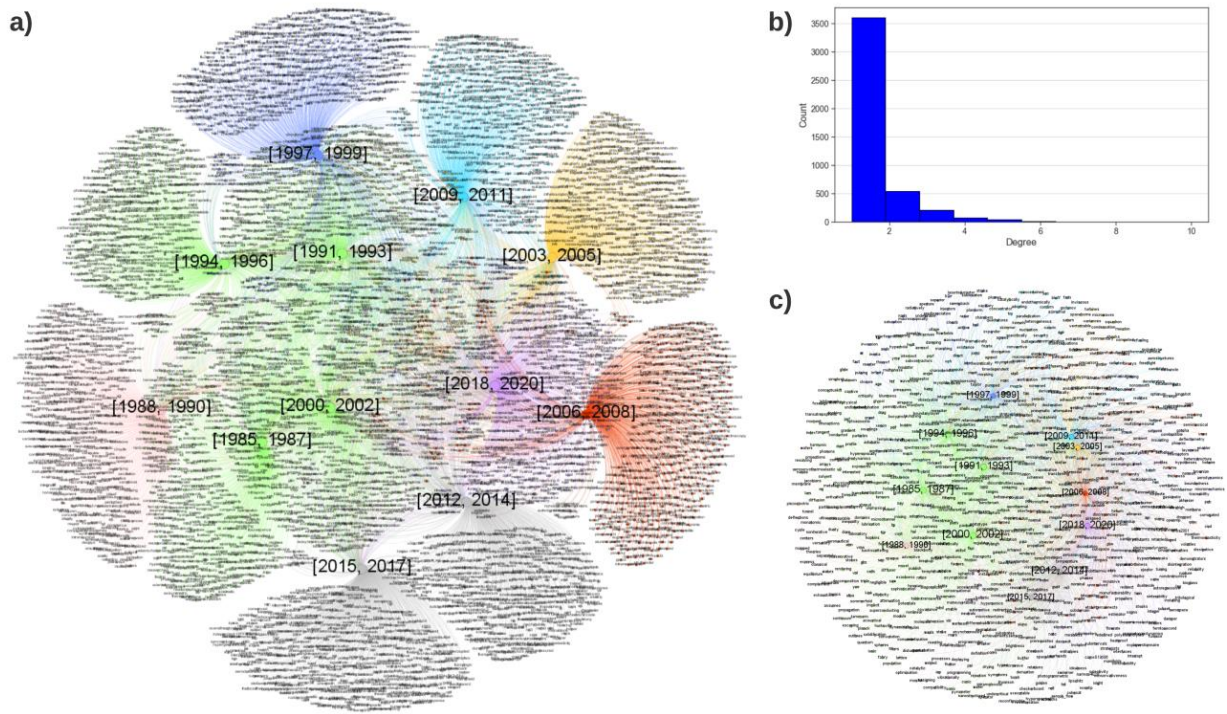

**Figure 6. a)** The <interval, keyword> network for the extracted term in t1 to t12, **b)** degree distribution of the terms, **c)** the <interval, keyword> network including terms with degree greater than one. Color in graphs represents network community.

After verifying the existence of a temporal pattern, we extract signals for each period. Table 2 lists alphabetically ordered the first 30 weak and strong signals per period, i.e., $P_1$, $P_2$, and $P_3$. Letters in brackets represent the category of the respective term. The table can be read at least in two ways: 1) the identified signals of the same category, i.e., weak/strong, can be compared to the same type of signals in other periods, and 2) different types of signals in the same period can be compared, e.g., weak vs. strong signals in $P_1$. This would provide insights on the evolution of the field, recent progress and/or focus, and oddities.

**Table 2.** The defined high-level categories of the extracted signals along with a selected set of keywords for each category.

| Period | Signal | Signal's keywords |
|---|---|---|
| $P_1$ | Weak (n = 112) | g]accelerometer, [v]acetylene, [a]aerohydrodynamic, a]aerophysical, [a]aeroshell, [v]afterburn, [m]anharmonicity, [m]b2o3, [g]backscattered, [m]beveled, [s]bhatnagar, [a]blast, ..., [m]throttling, [v]throughput, [a]tmk, [m]topologies, [g]tunability, [v]turbopump, [s]twin, [m]ultrarelativistic, [v]uncooled, [v]vehicular, [a]velocimetry, [a]windkanal |
| $P_1$ | Strong (n = 46) | [a]aerobrake, [m]aeroelastic, [a]aeronautics, [a]airfoil, [m]airframes, [s]burnett, [m]coat, [m]combustible, [m]combustors, [g]controllability, [v]coolant, [a]crossflow, ..., [s]schlieren, [v]scramjet, [g]sensors, [v]spacecrafts, [v]spaceplanes, [m]thermochemical, [a]thermomechanical, [v]thrust, [v]turbo, [a]vortex, [v]waveride, [g]yaw |
| P$_2$ | Weak (n = 153) | [v]acetylene, [a]aerogasdynamical, [a]aerohydrodynamic, [m]abstraction, [v]aegis, [a]aerophysical, [m]ageing, [m]airbag, [a]aeromechanics, [g]ailerons, [v]ammonia, [g]altimeter, ..., [m]vitrified, [s]vsl, [g]unguided, [v]vickers, [m]vulcan, [m]wafers, [a]waveguides, [g]waypoints, [g]whirl, [a]windside, [m]winglet, [m]zirconium |
| P$_2$ | Strong (n = 68) | [a]aerobrake, [m]aeroelastic, [a]aeronautics, [a]airfoil, [m]alloying, [a]aeroshell, [m]argon, [m]biconic, [s]bhatnagar, [a]blast, [v]capsule, [m]airframes, ..., [g]sensors, [s]reliability, [m]silicon, [v]thermocouples, [v]spacecrafts, [a]shockwave, [v]uavs, [m]topologies, [g]telemetry, [m]titan, [a]vaporization, [v]viscosities |
| P$_3$ | Weak (n = 131) | [v]acetylene, [a]aerohydrodynamic, [a]aeroph-flow, [m]airbag, [a]aerothermochemistry, [m]alkanes, [m]antiwindup, [v]atr, [v]autoignition, [m]b2o3, [g]backscattered, [s]bellman, ..., [v]turbomachinery, [v]turbopump, [v]turboramjets, [s]uhlenbeck, [m]underbody, [g]unguided, [s]vdf, [v]ventilation, [a]vibrometer, [m]viscoplastic, [m]wafers, [s]wcns |
| P$_3$ | Strong (n = 85) | [m]ablation, [m]abstraction, [g]aerodisks, [a]aeronautics, [a]aerospike, [m]alloying, [a]anisotropic, [v]apertures, [g]azimuthal, [v]ballistic, [g]boundedness, [v]bursts, ..., [a]streak, [a]swbli, [g]tangential, [v]tanks, [m]thermochemical, [v]throughput, [m]titan, [a]transmittance, [g]tunability, [g]vane, [a]vaporization, [m]zirconium |

Figure 7 depicts the temporal evolution of terms that are weak signals in either P$_1$ or P$_2$ and become strong signals in the subsequent period. Terms in black are the stems of the signals and the red terms in brackets are their corresponding terms. In the figure, "odw" stands for *Oblique Detonation Wave*, "pse" for *Parabolized Stability Equations*, and "swbli" for *Shock-Wave/Boundary-Layer Interaction*. The majority of these terms belong to the "Aerothermodynamics" and "Materials and structures" categories. Of course, some of the extracted signals are not useful by themselves to provide useful insights, e.g., "explosion" and "throughput". However, some terms are related to significant advancements in hypersonics over the last 30 years, e.g., "aeroshell", "magnetohydrodynamics", and "microstructures".

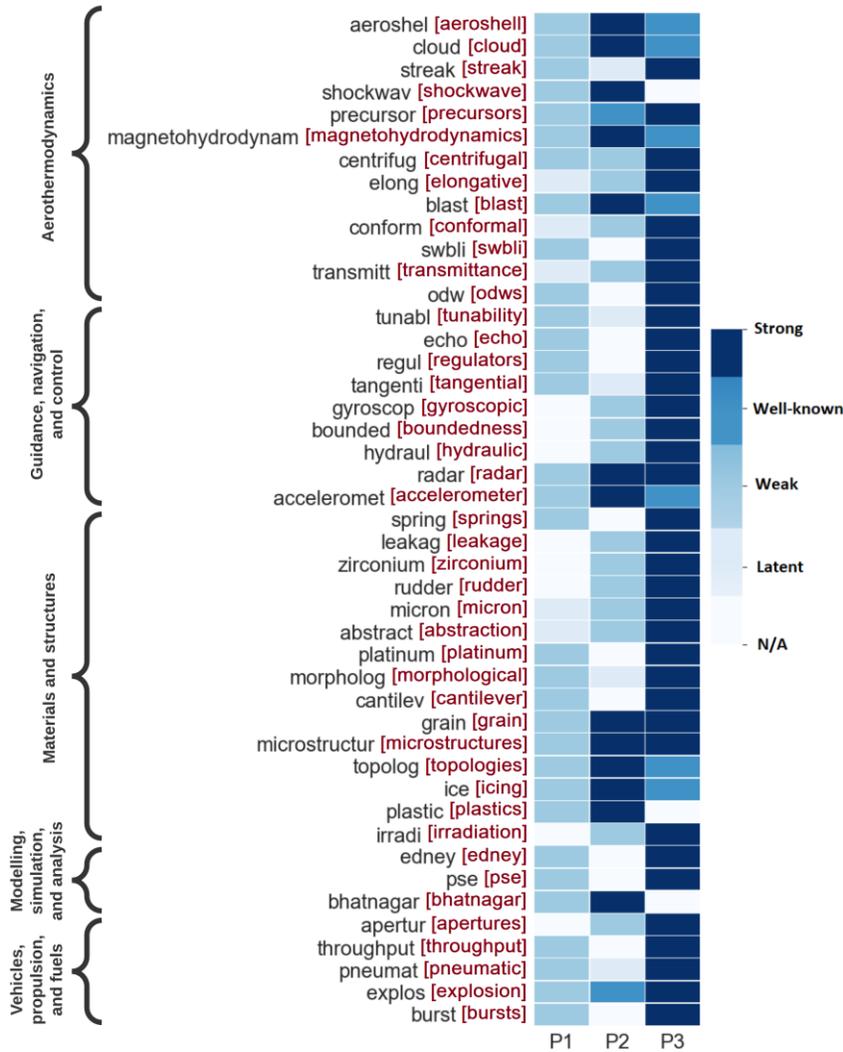

**Figure 7.** Weak signals in P$_1$ or P$_2$ that converted into a strong signal at P$_2$ or P$_3$. Terms in black are stems of the signals and the red terms in brackets are their corresponding terms. Stems are sorted based on their category.

Out of 131 weak signals in the final period (P$_3$), 58 of them have no appearance in the prior periods, suggesting they could be emerging weak signals of new research activities. Figure 8 lists these terms along with their representative category. The majority of the terms belong to the "Materials and structures" category, followed by "Aerothermodynamics" and "Modelling, simulation, and analysis".

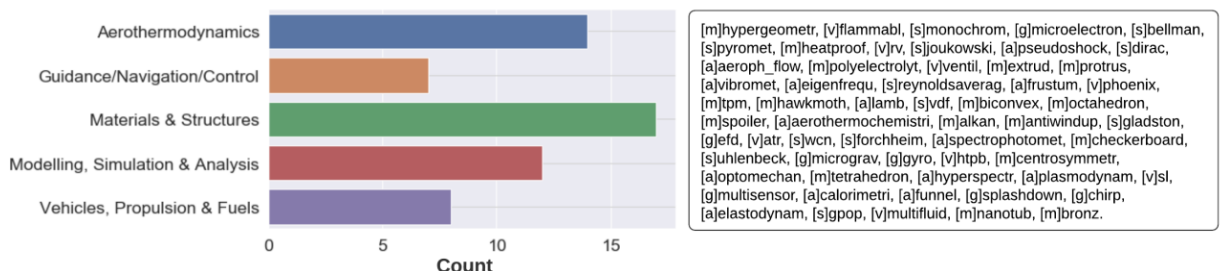

**Figure 8.** Weak signals in P$_3$ that did not appear as a signal in P$_1$ and P$_2$.

## 4.3. Terms with Specific Signal Patterns

The proposed framework allows strategic planners to extract terms with specific signal patterns of interest. Among the extracted signals, none of them follow a sinusoidal pattern over time, i.e., "*WeakP$_1$ → StrongP$_2$ → WeakP$_3$*" or "*StrongP$_1$ → WeakP$_2$ → StrongP$_3$*". This implicitly verifies the validity of the signals as sinusoidal patterns are not expected to be frequent in the evolving landscape of the case technology. However, some of the extracted signals follow a constant signal strength over time. "Controllability", "coolant", "fibers", and "decomposition" are strong signals in all the examined periods (n = 4), that may indicate enduring challenges mobilizing the research community. Similarly, other keywords remain weak signals in all periods of the dataset (n = 9). Examples include "mixer", "backscattered", "acetylene", "aerohydrodynamic", "brownian", and "striations".

## 5. Discussion and Conclusion

Given that the research field of hypersonics encompasses a wide range of technologies, thoroughly assessing the performance of weak signal detection is a challenging task. Nevertheless, we can still assess the usefulness of the approach by looking at which signals went from weak in P1 or P2 to strong later on. These terms are shown in Figure 7. Although they are all relevant to the field of hypersonics, many of them do not represent a technology per se but are rather indicators of scientific and technical research areas. Examples of such terms include "cloud", "ice", "abstraction", "tangential", or "echo". Another set of terms seems more specific to the measurement and observation of physical phenomena during experimentation than the technology itself, such as "blast", "burst", "explos", "shockwave", "swbli" (for shock-wave-boundary-layer interaction), and "throughput". Although these terms are not directly representative of a technology, they help understand how the field has been developing (e.g., engineers paying more attention to ice formation on wings) and still have value, especially when analyzed in the context of their category (e.g., propulsion). Furthermore, even for a broad field such as hypersonics, it is not too onerous for a subject-matter expert engaged in technology foresight to review the list and retain terms with signal patterns of interest.

In the present case, among the terms that remain, many represent actual technologies that had a major and demonstrable impact on the broader field of hypersonics[1], including:

- *Aeroshel* - Aeroshells are the rigid heat-shielded shells that help decelerate and protect spacecraft from pressure, with shapes now computationally designed to achieve specified lift-to-drag ratios [65].
- *Magnetohydrodynam* - Magnetohydrodynamic (MHD) bypass has become a common way to boost airbreather engine performance in hypersonic propulsion systems [63, 66]. The technology was originally used on Russian scramjet demonstrators.
- *Microstructur* - New thermal protection structure designs rely heavily on the conduct of complex aero-thermo-elastic simulations and thermoelastic materials [63].

Accordingly, we can conclude that for the particular field and dataset analyzed, the multi-layer approach presented in this paper has successfully identified weak signals associated with emerging technologies in the field of hypersonics, without presenting experts with an overwhelming number of technical terms to review and validate. This approach could thus have important benefits for organizations engaged in technology roadmapping as a means to inform and prioritize research and development activities.

---

[1] If we consider terms that went from *weak* to *well-known*, many also represent significant advances in the field, including *scramjet* engines that, although still under development, have been propelling many test aircraft already [63]. We also find *endothermic* reactions now leveraged to cool aircraft [63, 64], especially through "reformed" fuels such as *steam*-reforming hydrocarbon fuel that create endothermic reactions [63].

# 6. Limitations and Future Work

## 6.1. Limitations

As illustrated above, the approach often returns terms that, although relevant, do not seem directly associated with technologies (e.g., "hazard", "kw"), and as such the signal-to-noise ratio (i.e., precision) of the output could still be improved. Another limitation is that although the approach successfully identified early many technologies of importance in hypersonics, it also identified many of them very late. For example, the term "glider" was picked up as a *strong* signal in $P_3$, but not categorized at all in $P_1$ or $P_2$. Yet, some types of hypersonic platforms, such as boost-glide missiles, have been conceived over 80 years ago and in development since the early 2000s [67]. They are now becoming of concern due to claims of them being fielded in military operations [68]. This omission could be due to the use of a single source of data or to the fact that such platforms were linked to multiple stems (e.g., "boost", "glide") of different strengths. As such, this limitation might be more a limitation in the data or post-processing than a limitation of the approach itself. Nevertheless, it shows that the recall performance of the method could be improved to reduce the risk of missing important terms, but this would have to be done without compromising precision performance.

## 6.2. Future Work

A follow-on project will validate the proposed multi-layer approach by applying it to a different scientific and technical research area and re-assess the performance of the approach in detecting weak signals of emergence. The project will also consider the benefits of introducing new data sources beyond scientific publications, for instance, patent-related information.

# 7. Acknowledgements

We would like to thank Inbal Marcovitch and Karla Cisneros Rosado from Defence Research and Development Canada (DRDC) for their useful comments and suggestions. We also want to acknowledge the contribution of domain experts who reviewed and validated lists of signals.# References

[1] Y. LeCun, Y. Bengio, G. Hinton, Deep learning, nature 521 (7553) (2015) 436–444.

[2] S. Nakamoto, Bitcoin: A peer-to-peer electronic cash system, Decentralized Business Review (2008) 21260.

[3] U. Sahin, K. Karik´o, ¨O. T¨ureci, mRNA-based therapeutics—developing a new class of drugs, Nature reviews Drug discovery 13 (10) (2014) 759–780.

[4] X. Li, Q. Xie, L. Huang, Identifying the development trends of emerging technologies using patent analysis and web news data mining: the case of perovskite solar cell technology, IEEE Transactions on Engineering Management (2019).

[5] J. Rifkin, The third industrial revolution: how lateral power is transforming energy, the economy, and the world, Macmillan, 2011.

[6] T. S. Sechser, N. Narang, C. Talmadge, Emerging technologies and strategic stability in peacetime, crisis, and war, Journal of strategic studies 42 (6) (2019) 727–735.

[7] X. Li, Y. Zhou, L. Xue, L. Huang, Integrating bibliometrics and roadmapping methods: A case of dye-sensitized solar cell technology based industry in China, Technological Forecasting and Social Change 97 (2015) 205–222.14